\documentclass[letterpaper, 10 pt, conference]{ieeeconf}  

\IEEEoverridecommandlockouts                              

\usepackage{graphicx} 
\usepackage{amsmath} 
\usepackage{amssymb}  
\usepackage{cite}
\usepackage{color,soul}

\title{\LARGE \bf Reinforcement Learning Enabled Automatic Impedance Control of a Robotic Knee Prosthesis to Mimic the Intact Knee Motion in a Co-Adapting Environment}

\author{Ruofan Wu, Minhan Li, Zhikai Yao, Jennie Si, \textit{Fellow, IEEE} and He (Helen) Huang, \textit{Senior Member, IEEE}
\thanks{This work was partly supported by National Science Foundation: \#1563921 and \#1808752 for J. Si, \#1563454 and \#1808898 for H. Huang. Correspondence: J. Si and H. Huang.}
\thanks{R. Wu, Z. Yao and J. Si are with the the Department of Electrical, Computer,
	and Energy Engineering, Arizona State University, Tempe, AZ 85287 USA
	(e-mail: ruofanwu@asu.edu; zacyao.cn@gmail.com; si@asu.edu).}
\thanks{M. Li and H. Huang are with the Department of Biomedical Engineering,
	North Carolina State University, Raleigh, NC 27695 USA, and also
	with the University of North Carolina at Chapel Hill, Chapel Hill, NC 27599
	USA (e-mail:mli37@ncsu.edu;  hhuang11@ncsu.edu).}
}

\begin{document}

\maketitle
\thispagestyle{empty}
\pagestyle{empty}

\begin{abstract}

Automatically configuring a robotic prosthesis to fit its user's needs and physical conditions is a great technical challenge and a roadblock to the adoption of the technology. Previously, we have successfully developed reinforcement learning (RL) solutions toward addressing this issue. Yet, our designs were based on using a subjectively prescribed target motion profile for the robotic knee during level ground walking. This is not realistic for different users and for different locomotion tasks. In this study for the first time, we investigated the feasibility of RL enabled automatic configuration of impedance parameter settings for a robotic knee to mimic the intact knee motion in a co-adapting environment. We successfully achieved such tracking control by an online policy iteration. We demonstrated our results in both OpenSim simulations and two able-bodied (AB) subjects.

\end{abstract}

\begin{figure*}[!htp]
	\centering
	\includegraphics[width=400pt]{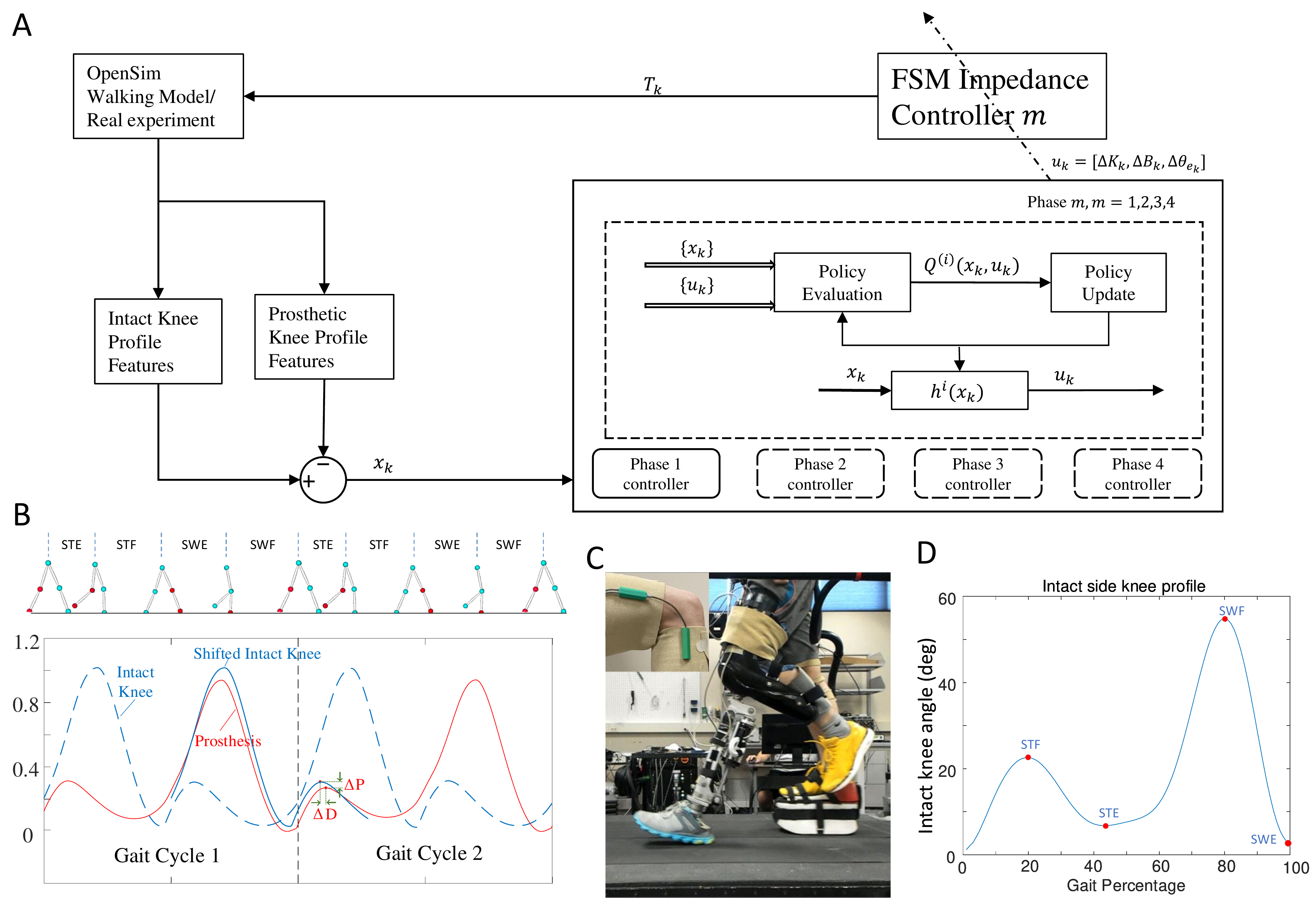}\\
	\caption{A) Schematic of RL tracking control of the robotic prosthetic knee using two human subjects. B) OpenSim simulation setup (Top). How the peak error and duration time error were obtained (bottom). Red: Prosthesis knee angle. Dashed Blue: Intact knee angle. Blue: intact knee angle shifted half cycle . C) Human testing environment. D) Feature extraction from the intact knee. It was based on the trajectory of the knee profile. Two peaks and troughs were used to determine the feature angles and phase transitions.}
	\label{fig:experiment}
\end{figure*} 

\section{INTRODUCTION}

Powered lower limb prosthesis as a new technological advancement has brought hope for amputees to regain mobility. Its potential has been demonstrated for transfemoral amputees’ walking ability on different terrains \cite{martinez2009agonist, sup2009self, johansson2005clinical, hafner2007evaluation}. State-of-the-art robotic device usually makes use of an impedance control framework which requires customization of the impedance parameters for each individual user. Currently, configuration of the powered devices is performed in clinics by technicians who manually tune a subset of impedance parameters over a number of visits of the patient.

Automatically configuring the impedance parameter settings has been attempted over the past several years.  An untested idea estimates the joint impedance based on biomechanical measurements and the model of the unimpaired leg \cite{ rouse2014estimation ,pfeifer2012model }. Another approach is to constrain the knee kinematics via the relationship of the joint control and intrinsic measurements \cite{ gregg2013experimental, eilenberg2010control}. A cyber expert system was proposed\cite{huang2016cyber} to simulate. It is not clear if the above approaches can become practical for use. 

We have successfully developed RL algorithms to configure impedance parameter settings and tested them in OpenSim simulations and in experiments using able-bodied and transfemoral amputee subjects \cite{ gao2018robotic, wen2018robotic, Li2019}. They are important milestones, yet have serious limitations. In all of our RL configuration approaches to date, the target knee motion profiles were all subjectively determined. As such, they are not feasible in realistic situations.

Mirroring the intact knee joint motion by a prosthetic knee is an intuitive solution as the intact knee kinematics is the most natural and realistic target: it contains actual biological joints’ inter-relational information, which makes it a good candidate to replace the subjective knee profile. It may be viewed as a virtual constraint that is directly measurable and thus avoid any modelling errors.

Studies have shown that bilateral coordination between two legs are needed in the regulation of bipedal walking to maintain stability, and also that such interlimb corporation can be accomplished at a spinal level. Since the spinal level locomotor network are symmetrically organized\cite{d2014modulation}, sensory and muscle activity of both sides are involved in rhythmic walking. Amputees usually display asymmetrical walking by relying heavily on their intact limb because of the loss of sensory feedback. Virtual constraints were proposed to generate coordinated joint motions as target joint motion profiles for the robotic knee to track\cite{kumar2019extremum}. Biomimetic virtual constraints described the joints’ geometric relationships and obtained a hybrid zero dynamics of the complete model \cite{ quintero2017toward, westervelt2003hybrid, westervelt2018feedback}. However, there are a few limitations on this approach. Virtual constraints require a simplified human model to establish the geometric relationship among joints. In a prosthesis control design based on the virtual constraints, only a proportional gain was derived and control performances are yet to be confirmed. 

Based on the above, mirroring the intact knee is an appealing approach, which has been explored years ago. Grimes et al. developed an mirror control scheme for the stance phase. It tracked the sound limb’s knee trajectory in the stance phase by multiplying a gain factor to avoid over flexion while a fixed trajectory was applied in swing phase. Melek et al copied the full gait trajectory but no human experiment was reported\cite{bernal2018design}. Joshi et al. developed a control strategy by controlling the swing time to mirroring the stride duration of the intact knee while the prothesis was locked during stance phase\cite{Joshi2010}. Sahoo et al aimed at mirroring the step length by controlling the push-off force\cite{sahoo2018novel}. The above approaches either focus on part of the gait cycle or the outcome measurement. None of them can track a completed gait cycle in practice. Tracking a complete gait cycle trajectory has not be achieved.

In this paper, we investigated the feasibility of mirroring the intact knee motion via reinforcement learning control. In our previous studies, we have successfully demonstrated RL regulation control to automatically configure impedance parameter settings while a subjectively prescribed knee motion profile is provided. Since RL is a data-driven, adaptive, and optimal control method, it is expected to be effective in solving the more challenging tracking control problem. We therefore applied an online policy iteration algorithm for the robotic knee to track a moving target (not the fixed, prescribed target knee motion anymore). In the following, we demonstrate results based on both OpenSim simulations and testing involving human subjects.

\section{Method}

This study was carried out using both OpenSim simulations and human subject experimental validation. The finite state machine (FSM) impedance controller (IC) was used as intrinsic controller with the impedance parameter settings automatically tuned to enable stable, continuous walking. 

\subsection{Finite State Machine (FSM) Impedance Control (IC)}
The FSM-IC is common for prosthesis intrinsic control as studies have shown that humans control the stiffness of leg muscles and therefore joint impedance while walking, and compliant behavior of legs are instrumental for human walking. The impedance controller generates a torque input (refer to equation\ref{equ:torque}) to the robotic knee based on current knee kinematics and knee joint impedance settings (equation \ref{equ:impedance}). 

Refer to Fig. \ref{fig:experiment}A, a gait cycle is divided into four phases in the FSM-IC: stance flexion (STF), stance extension (STE), swing flexion (SWF) and swing extension (SWE). The phase transitions are determined by knee motion and gait events (heel strike and toe-off) that are obtained from vertical ground reaction forces of both legs. In each phase of the FSM, three impedance parameters (stiffness $K$ damping $B$ and equilibrium position $\theta_e$) are provided as inputs to the FSM-IC for gait cycle $k$:
\begin{equation}
I_k = [K_k,B_k,\theta _{e_k}].
\label{equ:impedance}
\end{equation}
The knee joint torque is consequently generated by the following first principle equation
\begin{equation}
T_k=K_k(\theta-\theta_{e_k})+B_k\omega.
\label{equ:torque}
\end{equation}
The RL controller will adjust these parameters, i.e., 
\begin{equation}
u_k = [\Delta K_k,\Delta B_k,\Delta \theta _{e_k}].
\label{actioneq}
\end{equation}

so that updated impedance parameters are applied to the FSM-IC to generate knee torque
\begin{equation}
\begin{split}
K_{k+1} &= K_{k} +\Delta K_k, \\
B_{k+1} &= B_{k} +\Delta B_k, \\
\theta_{e_{k+1}} &= \theta_{e_{k}}+\Delta\theta_{e_{k}}. \\
\end{split}
\end{equation}
Therefore, we can write
\begin{equation}
I_{k+1} = I_k+u_k.
\end{equation}

\subsection{OpenSim Simulation}
We first investigated mirror control using OpenSim, a well-established simulator in the field of biomechanics \cite{delp2007opensim}, a bipedal walking model \cite{jacobs} that includes a body of five rigid-segments, linked through a one degree of freedom pin joint and the pelvis was linked to the ground by a free joint to allow free movement. The model settings, such as segment length, body mass and inertial parameters, followed the lower limb OpenSim model \cite{jacobs}. In this study, both knees and hips were controlled by FSM-IC controllers. With fixed impedance parameter settings while the prosthetic knee was controlled by the RL controller. Additionally, the second gait cycle of the prosthesis side was taken to extract the state features for the intact knee to establish the target profile for the prosthetic knee to track..

\subsection{Human Experiment Setup}
The experimental protocol was approved by the Institutional Review Board at the University of North Carolina at Chapel Hill. During the experiment, subjects wore a powered knee prosthesis and walked on a treadmill at a constant speed of their preferences as shown in Fig. \ref{fig:experiment}C. 

To acquire the intact knee kinematics, a goniometer provided by Biometrics Ltd. \cite{BiometricsLtd} was used to measure the knee angle (Fig. \ref{fig:experiment}C). The robotic knee prosthesis used in this experiment was designed based on \cite{Liu2014}. This prosthesis used a slider-crank mechanism, where the knee motion was driven by the rotation of the moment arm powered by the DC motor through the ball screw. An embedded potentiometer was used to record the robotic knee kinematics and an embedded load cell was used to trigger the phase transition. The prosthesis was controlled by a LabVIEW and MATLAB integrated system in a desktop PC with a 100 Hz sampling rate.

\begin{figure}[h]
	\centering
	\includegraphics[width=250pt]{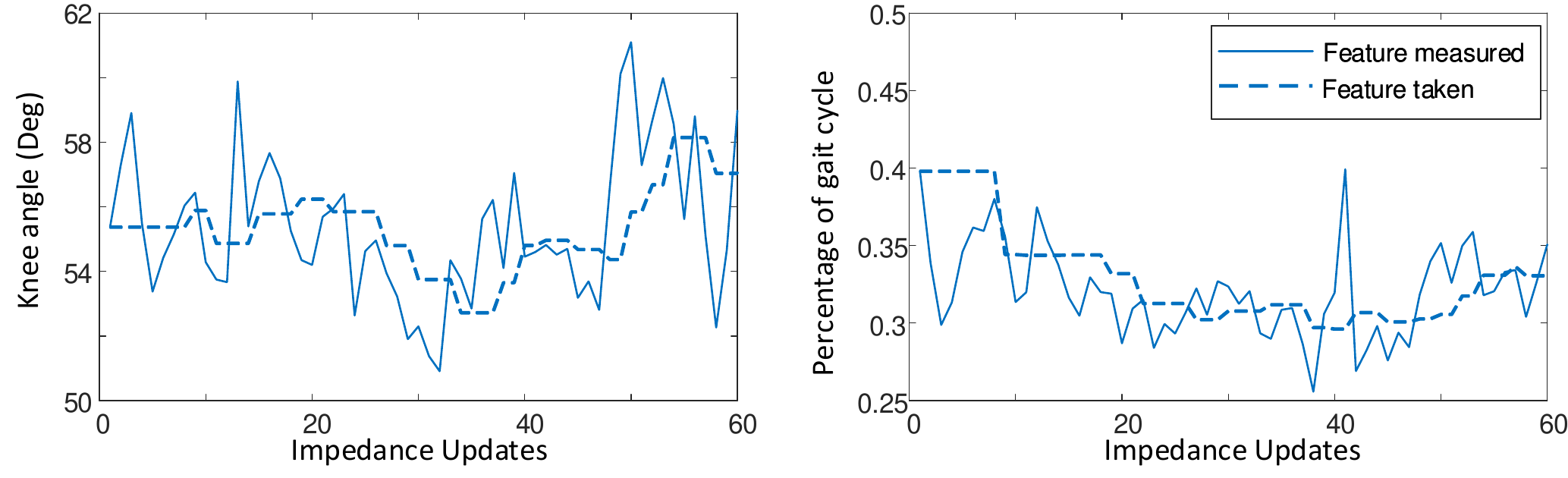}\\
	\caption{Target feature extraction from the intact knee for the robotic knee to mirror. Peak angle (Left) and phase duration (Right) are calculated as described in Section II D.}
	\label{fig:feature extration}
\end{figure}

\begin{figure*}[!ht]
	\centering
	\includegraphics[width=480pt]{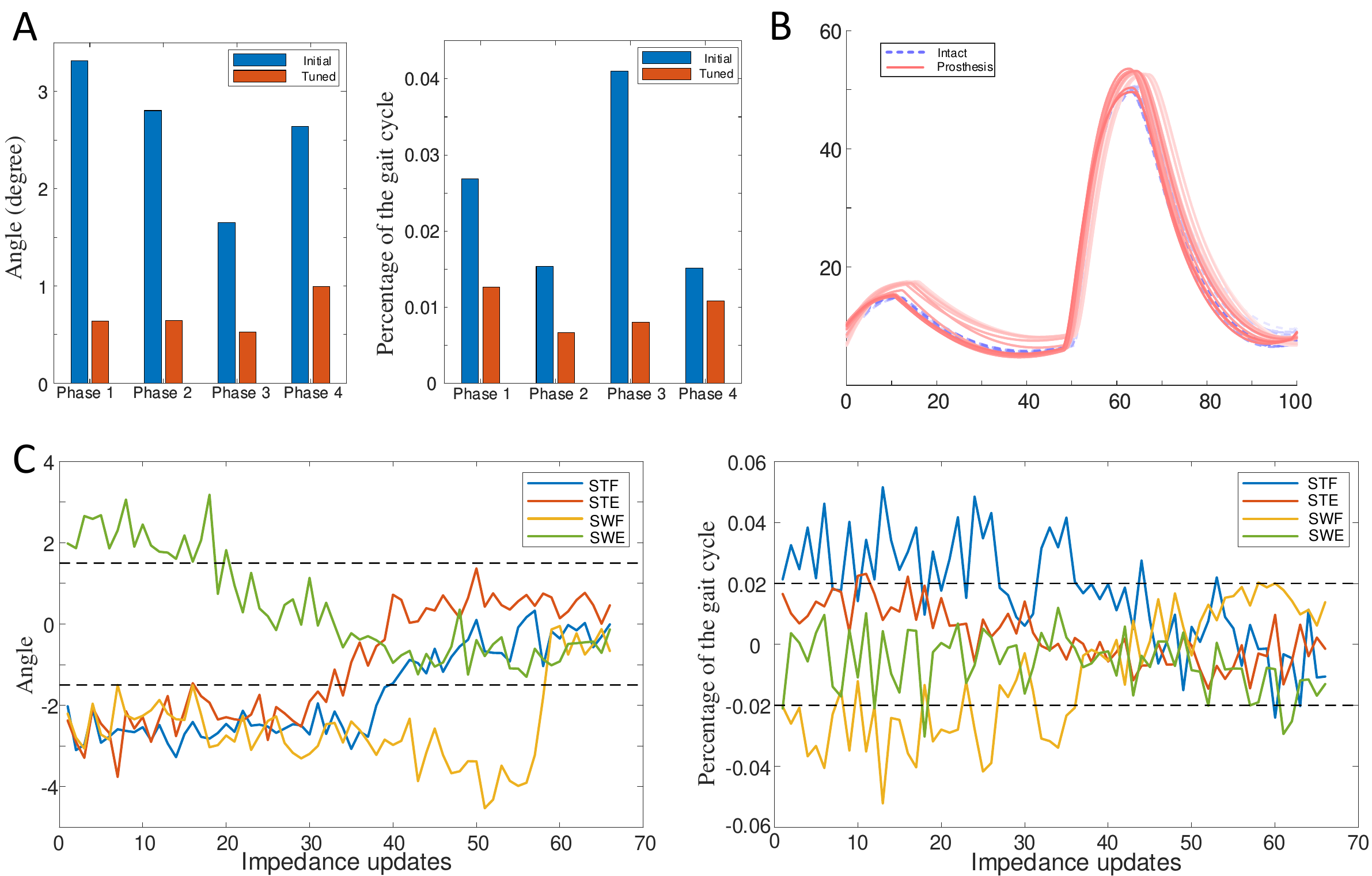}\\
	\caption{Results of OpenSim simulations. A) The RMSEs of peak angle error (left) and duration error (right) which were obtained from 30 trials. The initial errors (first 10 steps) and final errors (the last 10 steps) were represented, respectively. B) The peak angle error (left) and duration error (right) during an entire tuning process from one example trial was presented. C) Knee profiles of intact side and prosthesis in OpenSim simulations. Colors (from light to dark) indicates the order of phase tuning. Each trajectory was sampled every 5 impedance updates.}
	\label{fig:Opensim Result}
\end{figure*}

\section{Automatically Mirror the Intact Knee Motion by RL Tracking Control}

Fig.\ref{fig:experiment}A depicts the RL based solution approach to automatically configuring the impedance parameters of the robotic knee to mirror the intact knee joint motion. 

\subsection{Tracking Problem Formulation}
For a gait cycle index $k$, the intact knee motion featured by the peak knee angle $P^i_k$ (degrees) and duration $D^i_k$ (seconds) are measured. Similarly, we measure the peak knee angle $P^p_k$ and duration $D^p_k$ of the prosthesis (Fig. \ref{fig:experiment}B). Let $\Delta P_k$ and $\Delta D_k$ denote the peak value error and duration time error, respectively, i.e., 
\begin{equation}
\begin{split}
\Delta P_k &= P^p_k-P^i_k, \\
\Delta D_k &= D^p_k-D^i_k. 
\end{split}
\end{equation}

We have defined state $x_k$
\begin{equation}
x_k = [\Delta P_k,\Delta D_k ].
\label{StateEqu}
\end{equation}

We denote the RL state feedback control policy as
\begin{equation}
u_k=h(x_k).
\label{policy}
\end{equation}

Then we consider the instantaneous cost in a quadratic form 
\begin{equation}
U(x_k,u_k)={x_k}^TR_xx_k + {u_k}^TR_uu_k.
\label{policy1}
\end{equation}
where $R_x\in\mathbb{R}^{2\times2}$ and $R_u\in\mathbb{R}^{3\times3}$ are positive definite matrices.

As bilateral coordination between two legs are needed in the regulation of bipedal walking, the intact knee profile changes over time as the robotic knee adapts to the human user by tracking, specifically the intact knee motion. Thus, the tracking control problem is for RL control to enable mirroring trajectories $P^i_k$ and $D^i_k$  over $k$. As the intact knee motion as target trajectory changes over time, same for the human-robot interacting dynamics, it is nearly impossible to accurately describe them mathematically. As such, control theoretic approaches are inadequate to solve this tracking problem. Next, we formulate a RL based solution to solve this data-driven, tracking control problem. 

\subsection{Policy Iteration for Tracking Control}

We consider the tracking problem as one to devise an optimal control law via learning from observed data along the human-robot interacting system dynamics. For a control policy in (\ref{policy}), we define the state-action Q-function or the total cost-to-go as, 
\begin{equation}
Q\left(x_{k}, u_{k}\right)=U\left(x_{k}, u_{k}\right)+\sum_{j=1}^{\infty} U\left(x_{k+j}, h\left(x_{k+j}\right)\right).
\label{Q-function}
\end{equation}

Note that the $Q(x_k,u_k)$ value is a performance measure when action $u_k$ is applied at state $x_k$ and the control policy $h$ is followed thereafter. Such $Q(x_k,u_k)$ formulation implies that we have considered the optimal adaptive tracking control of the robotic knee as a discrete-time, infinite horizon, discounted problem without knowing an explicit mathematical description of the human-robot interacting dynamics.

Consider Q-value function in (\ref{Q-function}), it satisfies the Bellman equation
\begin{equation}
Q\left(x_{k}, u_{k}\right)=U\left(x_{k}, u_{k}\right)+Q\left(x_{k+1}, h\left(x_{k+1}\right)\right),
\end{equation}
An optimal control can be reached from solving the Bellman optimality equation,
\begin{equation}
Q^{*}\left(x_{k}, u_{k}\right)=U\left(x_{k}, u_{k}\right)+Q^{*}\left(x_{k+1}, h^{*}\left(x_{k+1}\right)\right).
\label{bellopti}
\end{equation}
where the optimal control policy $h^*$ is defined as
\begin{equation}
h^{*}\left(x_{k}\right)=\underset{u_{k}}{\arg \min } Q^{*}\left(x_{k}, u_{k}\right).
\label{equ:optimal policy}
\end{equation}
We applied a policy iteration procedure to solve equation (\ref{equ:optimal policy}). The policy iteration procedure starts from a random admissible initial policy $h^{(0)}$ and evolves from iteraton $k$ to $k+1$ based on the following policy iteration steps. Let $i$ denote the $i$th policy iteration step, policy iteration proceeds as follows.
\textit{ Policy Evaluation}
\begin{equation}
{Q}^{(i)}\left(x_{k}, u_{k}\right)=U\left(x_{k}, u_{k}\right)+{Q}^{(i)}\left(x_{k+1}, {h}^{(i)}\left(x_{k+1}\right)\right).
\label{policyEva}
\end{equation}
\textit{Policy Improvement}
\begin{equation}
{h}^{(i+1)}\left(x_{k}\right)=\underset{u_{k}}{\arg \min } {Q}^{(i)}\left(x_{k}, u_{k}\right), i=0,1,2, \ldots
\label{policyimp}.
\end{equation}

As analytical solutions to (\ref{policyEva}) and (\ref{policyimp}) are not available, we used the following approximation methods to seek solutions of Bellman optimality:
\begin{equation}
\hat{Q}^{(i)}\left(x_{k}, u_{k}\right)=W^{(i) T} \phi\left(x_{k},u_{k}\right)=\sum_{k=1}^{L} w^{(i)}_{k} \varphi_{k}\left(x_{k},u_k\right),
\label{equ:approximate}
\end{equation}
where $W^{(i)}\in\mathbb{R}^{L}$ is a weight vector and $\phi\left(x_{k},u_{k}\right): \mathbb{R}^{n} \times \mathbb{R}^{m} \rightarrow \mathbb{R}^{L}$ is a vector of the basis functions $\varphi_{k}\left(x_{k},u_k\right)$, which can be neural networks, polynomials, radial basis functions, etc. To prevent from a negative $\hat{Q}$ during approximation, a projection based modification to the policy iteration procedure was performed\cite{li2020towards}.
\begin{figure*}[!htp]
	\centering
	\includegraphics[width=480pt]{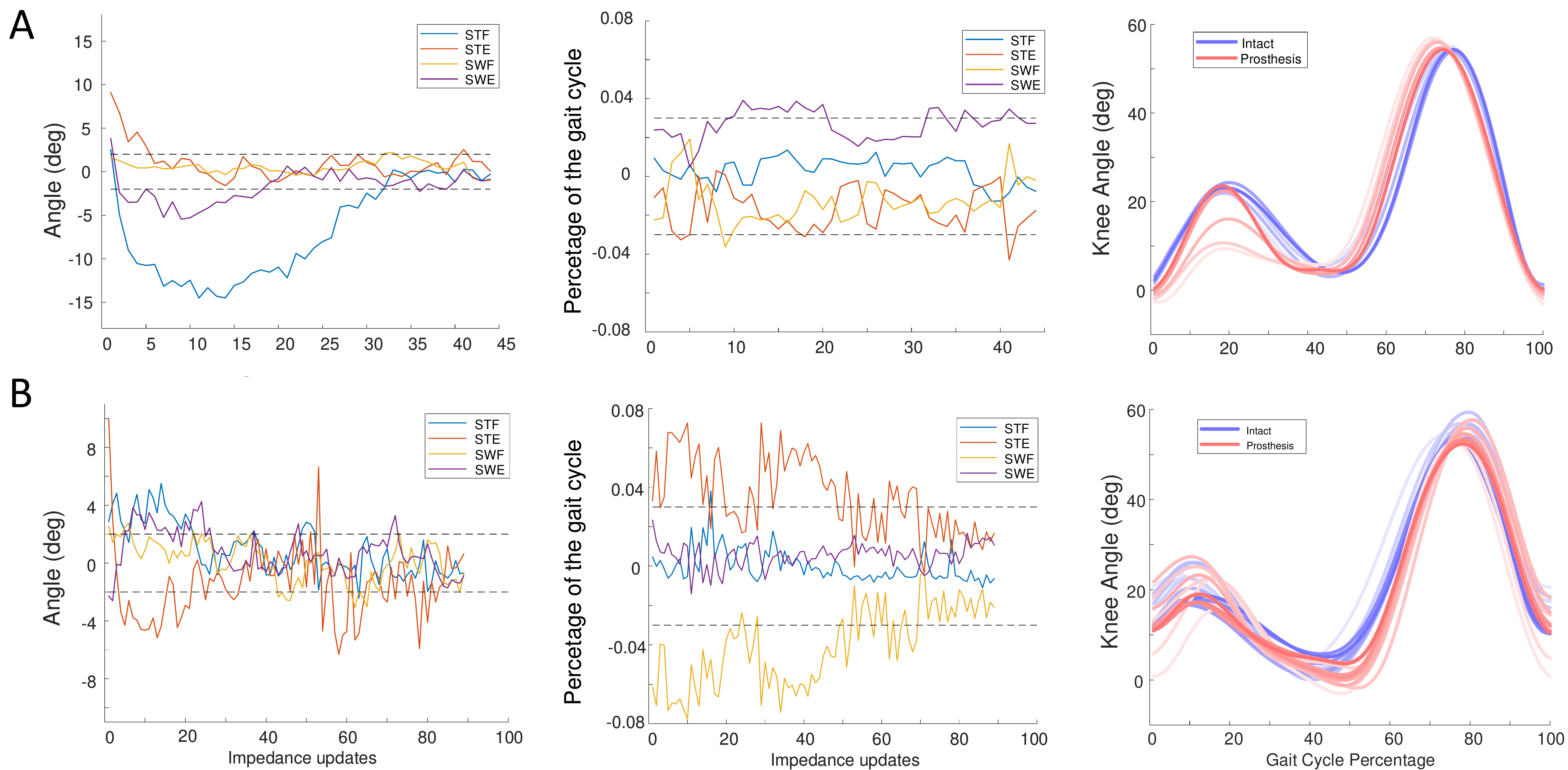}\\
	\caption{ Test results of two able-bodied subjects: A: AB1, B: AB2. The peak error (Left) and duration time error (middle) were presented, respectively. Knee profiles from both intact side and prosthesis of two subjects (Right). Colors (from light to dark) indicate the order of phase tuning. Each trajectory was averaged every 10 impedance updates.}
	\label{fig:FPP result}
\end{figure*}

The policy evaluation step (\ref{policyEva}) then becomes
\begin{equation}
\begin{array}{l}
\hat{Q}^{(i)}\left(x_{k}, u_{k}\right)
=U\left(x_{k}, u_{k}\right)+\hat{Q}^{(i)}\left(x_{k+1}, h^{(i)}\left(x_{k+1}\right)\right).
\end{array}
\label{equ:approeva}
\end{equation}
substituting (\ref{equ:approximate}) into (\ref{equ:approeva})
\begin{equation}
 W^{(i) T}\left(\phi\left(x_{k}, u_k\right)- \phi\left(x_{k+1}, h^{(i)}\left(x_{k+1}\right)\right)\right)=U\left(x_{k}, u_{k}\right).
 \label{wappro}
\end{equation}
The weight $W^(i)$ can be obtained by solving (\ref{wappro}).
\begin{equation}
W^{(i)}=X^{\dagger(i)}Y^{(i)}.
\label{equ:getw}
\end{equation}
where $X^{(i)} \in \mathbb{R}^{L\times N}$ and $y^{(i)} \in \mathbb{R}^N$ are two column vectors formed by $\phi\left(x_{k}, u_k\right)- \phi\left(x_{k+1}, h^{(i)}\left(x_{k+1}\right)\right)$ and $U\left(x_{k},u_{k}\right)$, respectively.

The $X^{\dagger(i)}$ is the Moore-Penrose pseudo-inverse of $X^{(i)}$. After $W^{(i)}$ is obtained from (\ref{equ:getw}), the policy improvement $h^{(i+1)}(x_k)$ can be obtained by
\begin{equation}
h^{(i+1)}\left(x_{k}\right)=\underset{u}{\arg \min } W^{(i) T} \phi\left(x_{k}, u_k\right).
\end{equation}
In obtaining the policy $h$ in (\ref{policy}), a second order polynomial basis was used.

\subsection{Algorithm Implementation}

For implementation of the tracking control, some details are needed. In OpenSim simulations, time index $k$ is a gait cycle while in human experiments, $k$ is every 4 gait cycles to obtain state features with good SNR. Policy iteration was initialized from a training session with a different subject using a policy iteration based approach\cite{li2020towards}.

Unlike the robotic knee motion that can be directly measured from embedded sensors on the prosthesis, the intact knee features (Fig. \ref{fig:experiment}D) has to be extracted from the knee joint motion profile to derive clear phase transitions. Four feature points corresponding to each of the four gait phases were extracted from the recorded profile which was measured by a goniometer. For the $k$th impedance update, the peak angle and duration from intact knee were obtained by applying a low pass filter of averaging 10 consecutive samples. A threshold of 1.5 degrees was also applied before updating to a new target feature value to further ensure robust estimate of the profile features. Fig. \ref{fig:feature extration} is an example of the intact knee feature.

\section{Results}
We implemented the RL control algorithm in both OpenSim and in real experiments involving two able-bodied subjects. OpenSim simulations allowed us to perform some systematic testing prior to human testing.

\subsection{Simulation Results}
Thirty (30) trials with random initial impedance (uniformly distributed with a variance at 0.2 of the initial impedance values) parameters were used in the reported results. A trial was convergent if it had reached a tracking error tolerance of 1.5 degrees for peak error and $2\%$ for duration error for 8 out of 10 consecutive impedance updates.
 
Fig.\ref{fig:Opensim Result}A is a summary of the simulation results based on 30 trials, which were all successful with an average tuning speed of $18.9$ tuning updates. Additionally, the root mean square error of peak errors was significantly decreased from $[3.31\ 2.80\ 1.65\ 2.64]$ to $[0.64\ 0.64\ 0.53\ 0.99]$ for all four phases while the duration error was decreased from $[2.7\%\ 1.5\% \ 4.1\%\ 1.5\%]$ to $[1.3\%\ 0.66\%\ 0.80\%\ 1.1\%]$, respectively. Fig. \ref{fig:Opensim Result}C is an example trial to illustrate the tuning process.

\subsection{Experiment Results}
Two able-bodied (AB) subjects participated in the human experiments. The two subjects walked at different speeds of their own preference, 0.65 m/s for AB1 and 0.7 m/s for AB2. The tracking error tolerance was 2 degrees for peak error and 3\% for duration error for 8 out of 10 consecutive impedance updates.

Fig. \ref{fig:FPP result} shows the peak and duration errors during tuning. The reduction of the errors of all four phases was observed from both subjects. For AB1, the peak error reduced from $10$ to $0.3$ degrees for STE and from $4$ to $-1$ degree for SWE. Interestingly, the peak errors of STF and SWE were initially within the convergence tolerance but quickly drifted out of bound. But the RL adaptation brought them back into tolerance range within 20 impedance updates. For AB2, the peak error of STE and STF were reduced from $10$ to $0$ and from $3$ to $-0.2$ degrees, respectively, while the other two phases initially were within tolerance and remained in the range. 

For the duration errors, AB1 was consistence during the test so that all four phases remained in error tolerance. In AB2, an apparent decrease during STE and SWF was observed from $6\%$ to $1.8\%$ and from $-6\%$ to $-2.5\%$, respectively, while the other 2 phases maintained a small phase duration error.

One-way ANOVA test was performed to verify significance of the results. The first 10 samples of the trial represent errors during the initial 10 tuning updates and the last 10 samples represent errors of the last 10 updates until convergence. The peak errors of AB1 decreased significantly ($p < 0.01$) while the duration time errors were not significant ($p > 0.01$) because it remained in the bound during test. For AB2, both peak errors and duration time errors decreased significantly ($p < 0.01$).

\section{Conclusions and Discussion}

This paper introduced a new RL based tracking control scheme for automatic tuning of robotic knee impedance
parameter settings for a robotic knee to mimic the intact knee kinematics. For the first time, we successfully demonstrated stable and continuous walking of a human wearing a robotic prosthesis which was controlled to automatically track the intact knee motion. We provided successful validation results of testing using OpenSim simulations and two human subjects.

Mirroring the intact knee motion by a prosthetic knee is an intuitive idea which has been proposed for decades, but have not been successfully demonstrated. The robotic knee control to mimic the intact knee joint is a tracking problem in classical control field. Even though many control theoretic solutions exist, such as backstepping\cite{krstic1995nonlinear}, observer-based control\cite{khalil2002nonlinear} and nonlinear adaptive/robust control\cite{nijmeijer1990nonlinear,isidori2013nonlinear}, they are inadequate for this problem as they require an accurate mathematical description of the system dynamics, which involve co-adapting human and robot in this case, and which are nearly impossible to obtain. Additionally, Those control theoretic approaches focus on the stabilization (in Lyapunov sense) of the nonlinear dynamic systems without addressing control performance such as convergence time and nice properties about transient dynamics.

Recently, some results emerged to tackle these issues using data-driven, learning enabled, nonlinear optimal tracking control designs\cite{hou2013model}.  Unfortunately, many of the reported results have focused on theoretical analyses, which are usually based on requiring a reference model for the desired movement trajectory and/or control trajectory. They are thus not practically useful.

An extremum seeking, continuous tracking idea was proposed very recently \cite{Kumar2017,kumar2019extremum}. This position controller uses virtual constraints of the limb joints to generate the target motion trajectory. The idea was applied to controlling knee and ankle joints by tuning a proportion gain controller. At this stage, there appears some limitations. The restoration of the desired trajectory from the virtual constraint needs more validation and justification, the proportional gain control is yet to be validated for accurate tracking performance either by simulations or by human testing.

In this study, the initial policy was transferred from a training session of a different subject. We had started testing using random initial polices without any transfer and were successful with the initial trials. But unfortunately, testing abruptly stopped due to coronavirus caused closure of the laboratory. We will report this result once systematic testing completes.

For the first time, we recorded the intact knee motion as the intact knee adapted to cooperate with the prosthesis to enable walking. Interestingly, the intact knee profile changed significantly for both test subjects (Fig. \ref{fig:FPP result}, the right column). Each line in Fig. \ref{fig:FPP result} was an average of 10 impedance updates. The lighter lines represent beginning of the tuning and the darker lines represent after tuning. It is apparent that the intact knee angle profiles changed during the tuning process which means humans adapted their intact body motion to cope with the prosthetic knee.  While this is helpful to further understand how using a prosthesis may influence an amputee’s gait pattern, the two tested subjects showed different adaptation patterns. AB1 tends to increase flexion in the STF phase but AB2 tends to reduce flexion. This may suggest that individuals use different approaches to collaborating with a prosthesis. 

In a previous work, Grimes \cite{Grimes1983} suggested that the intact leg may provide compensation when using a prosthesis by an over-flexion in the stance. We observed similar compensation behavior in the experiment. However, in our experiments, flexion was still within the safe range and the subjects didn’t feel uncomfortable during walking. AB2 even mitigated over-flexion as tuning proceeded. In addition, we observed that the intact knee was not fully extended during swing, which may cause over-flexion in the stance phase. One explanation is that the subjects wore a high-shoe on the intact knee to adjust the length difference of the able-bodied prosthesis socket. The high shoe may change the geometry of the intact knee foot which may cause an abnormal heel-strike. We may not see this happen on amputee subjects.

Our new results of RL enabled tracking control provides a great opportunity to explore the potential benefit of a symmetrical gait. Previous research revealed health benefits of a symmetrical walking pattern \cite{Kaufman2012}. Although our human experiment result did not show a consistent improvement in the temporal symmetry calculated by ground reaction force measures, the OpenSim simulation indicates a great potential for improving symmetrical walking. As only temporal symmetry was measured in the experiment, which limited our analysis. it is worth a further study on the symmetry including loading, step duration and step length. The natural asymmetry of the able-bodied participants who wore a high-shoe could explain the temporal asymmetry in human experiment result. Also, the tracking error tolerance was set to 3\% for the duration time, which can accrue phase by phase. Further study involving more individuals with lower limb amputations are needed to provide statistical test results on both temporal and spatial symmetry under mirror control.

\bibliographystyle{IEEEtran}

\bibliography{ICRA}

\end{document}